\documentclass{esannV2}
\usepackage{graphicx}
\usepackage[latin1]{inputenc}
\usepackage{amssymb,amsmath,array}
 \usepackage{url} 
 
 \usepackage{listings}
\usepackage{float}
\usepackage{placeins}

\newcommand{\R}{\mathbb{R}}

\newcommand{\y}{\textbf{y}}
\renewcommand{\u}{\textbf{u}}
\newcommand{\x}{\textbf{x}}
\newcommand{\W}{\textbf{W}}
\newcommand{\Win}{\W_{in}}

\newcommand{\Wout}{\W_{out}}

\begin{document}
\title{Deep Echo State Networks for Diagnosis of Parkinson's Disease}

\author{Claudio Gallicchio, Alessio Micheli and Luca Pedrelli
\vspace{.3cm}\\
Department of Computer Science, University of Pisa\\
Largo Bruno Pontecorvo 3 - 56127 Pisa, Italy\\
}

\maketitle

\begin{abstract}
In this paper, we introduce a novel approach for diagnosis of Parkinson's Disease (PD) based on deep Echo State Networks (\mbox{ESNs}).
The identification of PD is performed by analyzing the whole time-series collected from a tablet device during the sketching of spiral tests, without the need for feature extraction and data preprocessing.
We evaluated the proposed approach on a public dataset of spiral tests. 
The results of experimental analysis show that DeepESNs 
perform
significantly better than shallow ESN model.
Overall, the proposed approach obtains state-of-the-art results in the identification of PD on this kind of temporal data.

\end{abstract}

\section{Introduction}
Parkinson's disease (PD) is a neurodegenerative disease that mainly affects the extrapiramidal motor system causing 
tremor, bradykinesia, rigidity and loss of postural reflexes \cite{jankovic2008parkinson}.
The analysis of motor capacities such as handwriting and sketching abilities of patients is performed to assess and diagnose PD \cite{saunders2008validity}. While handwriting abilities are influenced by language capacities, sketching abilities involved in the execution of a spiral test (i.e., drawing a spiral with a pen) are considered as measures independent of education \cite{saunders2008validity}. An example of application of deep learning models for PD classification, applied directly to images of spiral tests, is introduced in \cite{pereira2016convolutional}.
In \cite{tablet} it is introduced a study on PD diagnosis 
through statistical methods
based on the analysis of velocity and the pen pressure data collected from a tablet device during the execution of the spiral test. 

In this paper, we propose a novel approach for diagnosis of PD based on Recurrent Neural Networks (RNNs) that are able to robustly exploit the whole time course of noisy and heterogeneous time-series data
collected from a tablet device during the execution of spiral tests. In particular, we consider the Reservoir Computing (RC) \cite{Jaeger2007,verstraeten2007experimental} framework that obtained state-of-the-art results in clinical assessment \cite{bacciu2017learning} related to the study of neurological diseases. In the application presented in \cite{bacciu2017learning}, signals characterized by noisy and heterogeneous time-series are analyzed for postural stability assessment by using Echo State Networks (ESNs). Recently, the introduction of DeepESN model as the extension of the Reservoir Computing (RC) \cite{Jaeger2007,verstraeten2007experimental} paradigm to the deep learning framework allowed to study the intrinsic properties of stacked RNNs \cite{gallicchio2017echo}
and to build efficiently trained models able to develop a hierarchical temporal
representation of the input signals, resulting in a suitable method to deal with tasks featured by multiple time-scales dynamics
\cite{gallicchio2017deep, gallicchio2017hierarchical}. Based on such considerations, we 
think 
that the DeepESN model can be particularly suitable for the study of PD through the analysis of tablet signals.

We assess
our 
approach for diagnosis of PD on a publicly available dataset of spiral tests introduced in \cite{isenkul2014improved}. Moreover, in order to investigate the practical relevance of using layered recurrent architectures in this domain, the experimental analysis is also conducted in comparison to the standard ESN. Finally, we 
evaluate the results achieved by ensembling the realizations of the selected model,  
in order to improve the classification performance of the proposed approach.

\section{Deep Echo State Networks}
\label{sec.DeepESN}
\begin{figure}[b]
\center
    \includegraphics[width=1\textwidth]{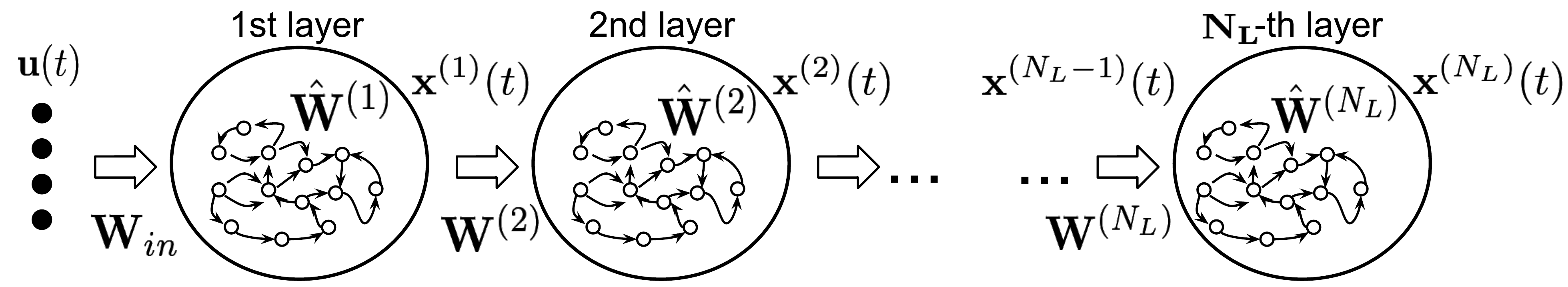}
    \caption{The hierarchy of reservoirs in the architecture of a DeepESN.}
\label{fig:DeepESN}
\end{figure}

DeepESNs 
\cite{gallicchio2017deep}
extend Leaky Integrator ESNs (LI-ESNs)
\cite{Jaeger2007} 
to the deep learning framework. 
A DeepESN architecture (see Fig. \ref{fig:DeepESN} for an example) 
is composed by a hierarchy of $N_L$ reservoirs.
Here we denote by $\u(t) \in \R^{N_U}$ 
and $\x^{(l)}(t) \in \R^{N_R}$, respectively, the input and state of the l-th reservoir layer
at step $t$.
Avoiding the bias term for the ease of notation, the state transition function of the first layer is computed by the following equation:
\begin{equation}
\label{eq.layer1}
\x^{(1)}(t) = (1-a^{(1)}) \x^{(1)}(t-1) + a^{(1)} \mathbf{tanh}(\Win \u(t) + \hat{\W}^{(1)} \x^{(1)}(t-1)),
\end{equation}
while for higher 
layers $l>1$, the state transition function is described as follows:
\begin{equation}
\label{eq.layeri}
\x^{(l)}(t) = (1-a^{(l)}) \x^{(l)}(t-1) + a^{(l)} \mathbf{tanh}(\W^{(l)} \x^{(l-1)}(t) + \hat{\W}^{(l)} \x^{(l)}(t-1)).
\end{equation}
In the above eq.~\ref{eq.layer1} and \ref{eq.layeri},
$\Win \in \R^{N_R \times N_U}$ is the matrix of input weights, $\hat{\W}^{(l)} \in \R^{N_R \times N_R}$ is the matrix of the recurrent weights of the layer $l$, $\W^{(l)} \in \R^{N_R \times N_R}$ is the matrix of weights relative to the inter-layer connections from layer $l-1$ to layer $l$, $a^{(l)}$ is the leaky parameter at layer $l$ and $\mathbf{tanh}$ is the activation function of recurrent units implemented by a hyperbolic tangent. Finally, the state of the DeepESN is given by the concatenation of all the states encoded in the recurrent layers of the architecture $\x(t) = (\x^{(1)}(t),
\ldots, \x^{(N_L)}(t)) \in \R^{N_L N_R}$.

Let $\sigma$ and $\hat{\sigma}$ be the scaling of the external input and the inter-layer scaling, then the weights in matrices $\W_{in}$ and $\{\W^{(l)}\}_{l = 2}^{N_L}$ are randomly initialized from a uniform distribution and re-scaled such that $\lVert \W_{in} \rVert{_2} = \sigma$ and $\lVert \W^{(l)} \rVert{_2} = \hat{\sigma}$ respectively. According to the necessary condition of Echo State Property for DeepESNs \cite{gallicchio2017echo}, values of matrices $\{\hat{\W}^{(l)}\}_{l=1}^{N_L}$ are randomly initialized from uniform distribution and rescaled such that $\max\limits_{1 \leq l \leq N_L} \rho\left((1-a^{(l)}) \mathbf{I} + a^{(l)}\hat{\W}^{(l)}\right) < 1$, where $\rho$ denote the spectral radius of the matrix (i.e., the largest absolute value of its eigenvalues). Note that considering a number of layers $N_L=1$ the recurrent architecture described above is equivalent to a LI-ESN, 
that is referred to as
shallowESN in the following.

For the classification task considered in this work, the DeepESN computes an output in correspondence of an entire input sequence. 
Accordingly, we compute a mean state mapping as follows: $\chi(\mathbf{s}) = \frac{1}{n}\sum_{t=1}^n \mathbf{x}(t),$ 
where $\mathbf{s} = [\mathbf{u}(1), ..., \mathbf{u}(n)]$ is an input sequence of length $n$.
Then, the output of the network
is computed as
$\y(\mathbf{s}) = \Wout \chi(\mathbf{s})$, where $\Wout \in \R^{N_Y \times N_L N_R}$ is the
weights matrix relative to the readout layer. The linear combination between the states encoded in the recurrent layers and the weights of the readout allows to differently weight the contribution of the multiple dynamics developed in the network state.
As in the standard RC framework, only the readout layer is trained, typically by means of direct method such as pseudo-inversion or ridge regression \cite{Jaeger2007,verstraeten2007experimental}.

\section{DeepESNs for Diagnosis of PD through Spiral Tests}
\begin{figure}[h]
\centering
\includegraphics[width=0.45\linewidth]{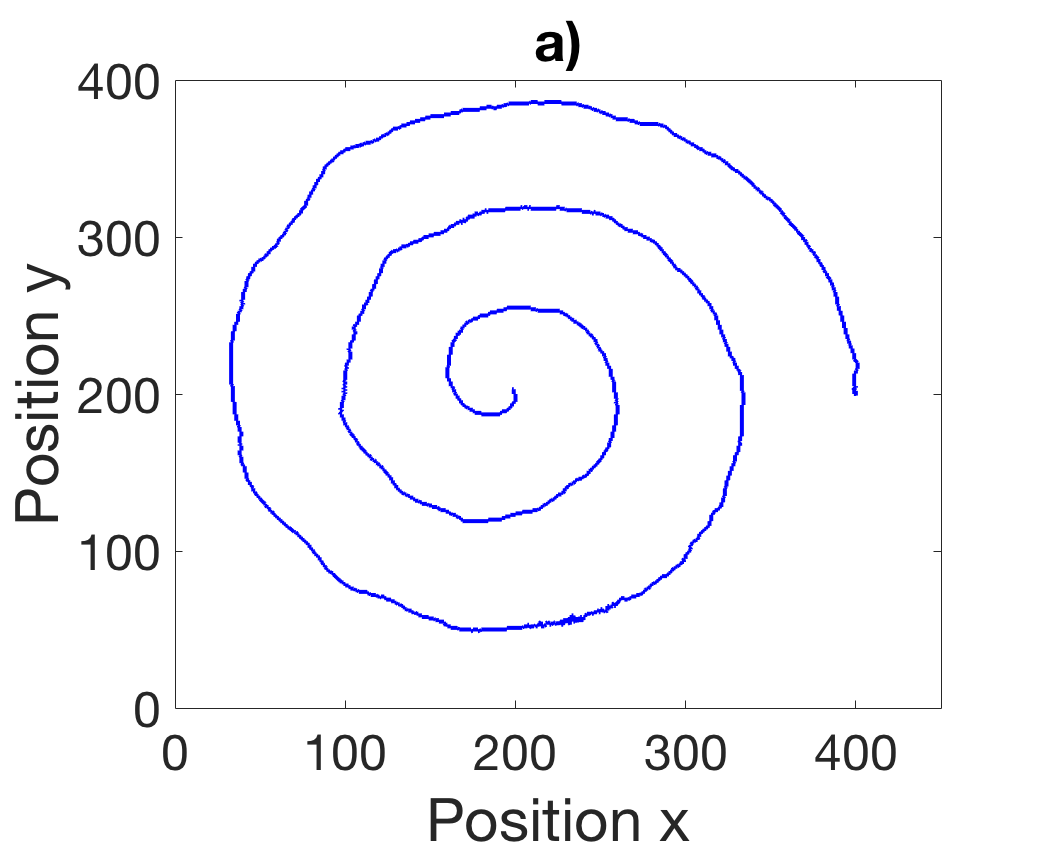}
\includegraphics[width=0.45\linewidth]{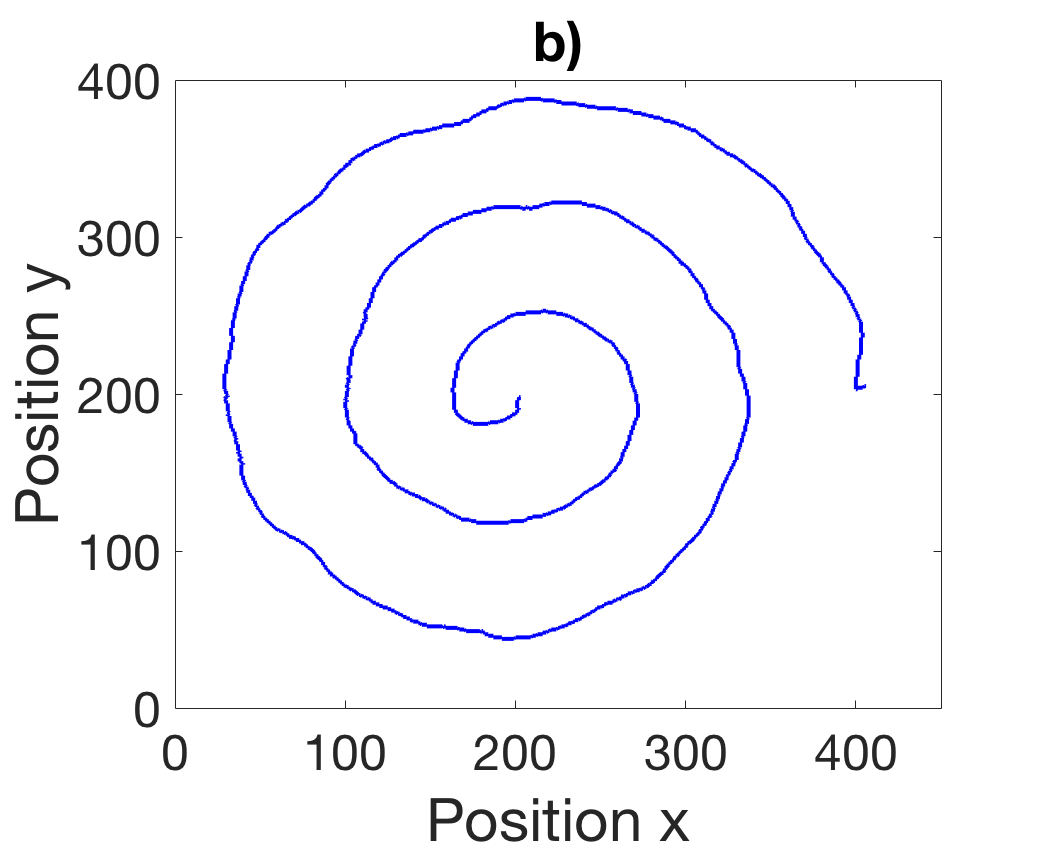}
\includegraphics[width=0.45\linewidth]{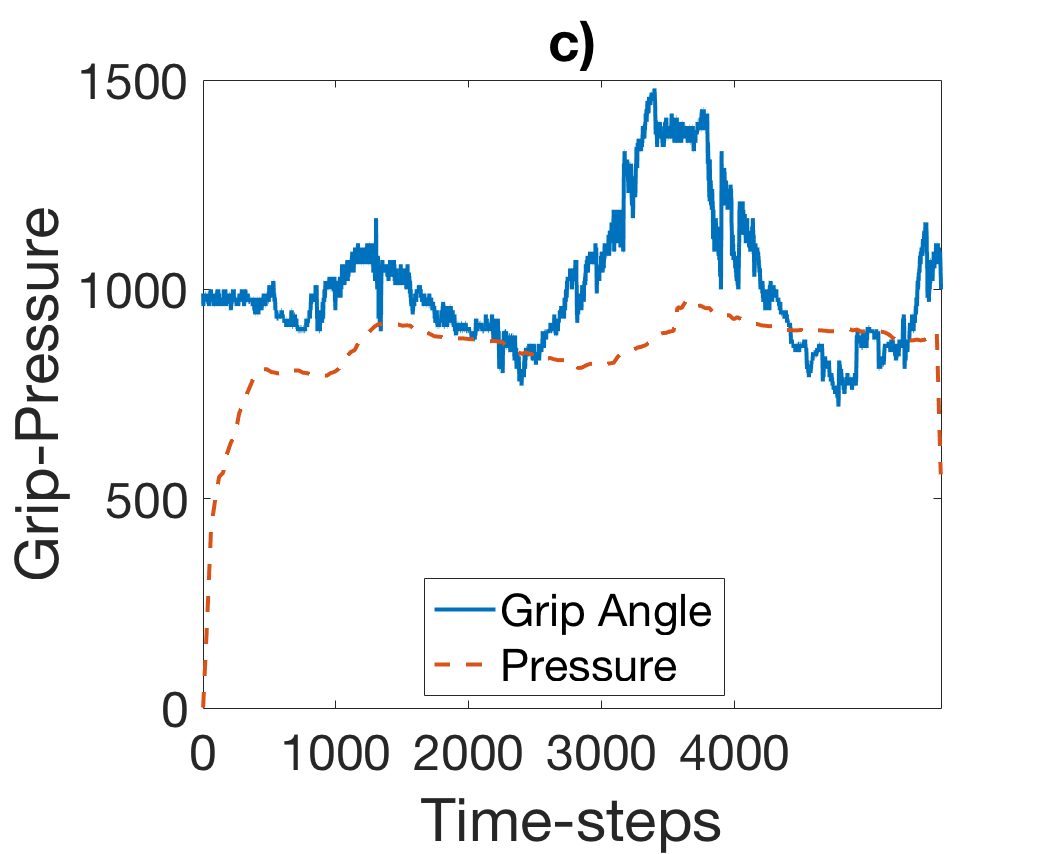}
\includegraphics[width=0.45\linewidth]{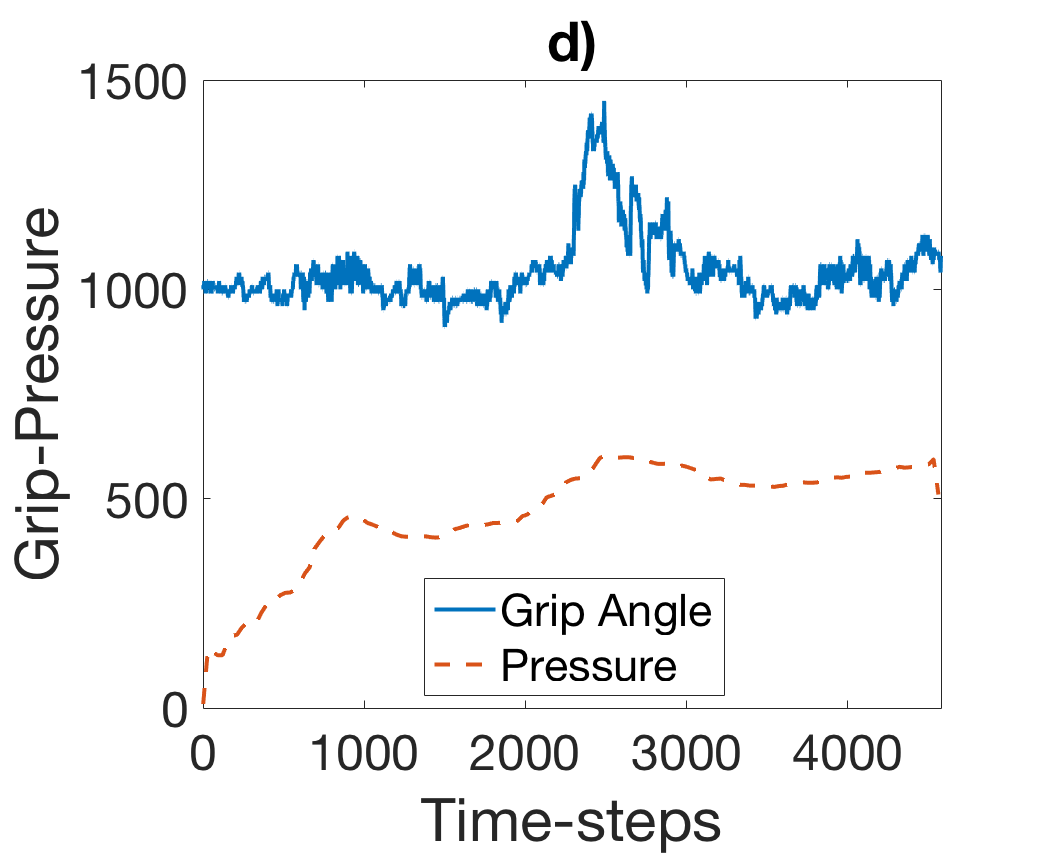}

\caption{Pen positions (1st row),
pressure and grip angle (2nd row) 
gathered for 
a control (fig. a), c)) and a PD (fig. b), d)) patient.}
\label{fig:spiral_test}
\end{figure}

We evaluated the proposed DeepESN model on the spiral dataset described in \cite{isenkul2014improved}. This dataset is publicly available on the UC Irvine Machine Learning Repository\footnote{\url{<https://archive.ics.uci.edu/ml/datasets/Parkinson+Disease+Spiral+Drawings+Using+Digitized+Graphics+Tablet>}}. The dataset is composed by spiral tests executed on a tablet device by 61 PD patients and 15 control patients without PD. For each time-step, the time-series gathered from the tablet contain pen position (x and y components), pen pressure and grip angle. In our method, the models were fed directly with such 
signals
without feature extraction or data preprocessing. 
Note that, the whole 
temporal
signal 
generated by the tablet potentially contains a richer information than static preprocessed features, 
thereby a RNN model can be
more effective on the analysis of this kind of data. Fig. \ref{fig:spiral_test} a), \ref{fig:spiral_test} b), \ref{fig:spiral_test} c) and \ref{fig:spiral_test} d) show (x,y) pen position, pen pressure and grip angle gathered at each time-step from tablet during the execution of the test by a control (Fig. \ref{fig:spiral_test} a), \ref{fig:spiral_test} c)) and a PD (Fig. \ref{fig:spiral_test} b), \ref{fig:spiral_test} d)) patient. As we can note, by visual inspection the sketches showed in Fig. \ref{fig:spiral_test} a) and \ref{fig:spiral_test} b) are quite similar and consequently the classification task is not trivial in such cases. 
Moreover, from Fig. \ref{fig:spiral_test} c) and \ref{fig:spiral_test} d) it is possible to see that the signals relative to pen pressure and grip angle are rather noisy, therefore, the analysis of such kind of data without feature extraction can be challenging since we need a noise-robust approach to perform a correct classification.

We evaluated the generalized performance of the proposed approach through a 3-fold cross validation, inserting in each fold 20 (or 21) PD and 5 control patients. We considered a 10-layered DeepESN (i.e. $N_L = 10$)
with the same number of $N_R$ units per layer  and a leaky integrator of $a^{(l)} = 0.1$ for each layer $l$.
The rest of the DeepESN hyper-parametrization were selected from the ranges reported in Tab. \ref{tab:hyper_parameters}, on a validation set by an extra level of 5-fold validation on each fold. For each hyper-parametrization, we generated 10 reservoir guesses, averaging the results on such guesses.
\begin{table}[h!]
  \centering
\begin{center}
\begin{tabular}{|l|l|}
\hline
 \textbf{Hyper-parameter} & \\
\hline
recurrent units per layer $N_R$ & $10, 20, 30, 40, 50$ \\
\hline
input scaling $\sigma$ & 0.1, 0.5, 1, 2 \\
\hline
inter-layer scaling $\hat{\sigma}$ & 0.1, 0.5, 1, 2 \\
\hline
spectral radius $\rho$ & 0.7, 0.8, 0.9, 1.0 \\
\hline
readout regularization $\lambda_r$ & $0, 10^{-10}, 10^{-9}, ..., 10^0$ \\
\hline
\end{tabular}
\end{center}
\caption{Range of DeepESN hyper-parameters values for model selection.}
\label{tab:hyper_parameters}
\end{table}

Tab. \ref{tab1} shows the accuracies and the standard deviations on reservoir guesses (in parenthesis) obtained by DeepESN and shallowESN in PD classification task. In order to perform a fair comparison, we selected the shallowESN (i.e., 1-layered DeepESN) considering the same hyper-parameters ranges
as for 
DeepESN and the same range of total number of recurrent units
(i.e. 100-500), where the 
specific choice depends on the
fold of the 3-fold cross validation.
\begin{table}[h!]
  \centering
  \begin{tabular}{|l|l|l|l|}
    \hline
    Model & TR & VL & TS\\
    \hline
    DeepESN & \textbf{94.27\% (1.18\%)} & \textbf{86.57\% (2.62\%)} & \textbf{87.20\% (2.78\%)} 
    \\
    shallowESN & 91.60\% (0.56\%) & 83.62\% (3.39\%) & 84.13\% (2.63\%) \\     
    \hline
  \end{tabular}
  \caption{Accuracies and standard deviations on reservoir guesses (in parenthesis) obtained by DeepESN and shallowESN in validation, training and test.}
  \label{tab1}
\end{table}
Interestingly, DeepESN model outperformed shallowESN in training, validation and test set of $2.67\%$, $2.95\%$ and $3.07\%$ of accuracy respectively. This comparison result suggests that the ability of DeepESN in providing a hierarchical temporal representation of input signals with complex dynamics allows us to improve the performance on this kind of tasks. For the sake of completeness, we statistically compared the accuracies obtained on test set by DeepESN and shallowESN models through McNemar's test. The classification accuracies obtained by DeepESN and shallowESN resulted significantly different with a p-value of 0.0032.

To exploit the variability provided by different reservoir guesses, we evaluated the selected model considering 
an
ensemble approach. Accordingly, the classification was performed averaging the output of the different guesses of the selected model.
Results are reported in Tab.~\ref{tab2}, and show that the ensemble approach
allowed DeepESN to achieve improved performance in training, validation and test set of $0.40\%$, $2.95\%$ and $2.13\%$ of accuracy respectively. Overall, our proposed automatic system obtained a test accuracy of $89.33\%$ with a sensitivity (percentage of PDs correctly classified) and a specificity (percentage of controls correctly classified) in test set of $90.00\%$ and $80.00\%$ respectively. 

\begin{table}[htb]
  \centering
  \begin{tabular}{|l|l|l|l|l|l|}
    \hline
    Model & TR ACC & VL ACC & TS ACC & TS SEN & TS SPEC\\
    \hline
    DeepESN&94.67\%&89.52\%&\textbf{89.33\%}&\textbf{90.00\%}&\textbf{80.00\%}\\    
    \hline
  \end{tabular}
  \caption{Accuracy (ACC), sensitivity (SEN) and specificity (SPEC) obtained by ensemble of DeepESN 
  in training (TR), validation (VL) and test (TS) set.
  }
  \label{tab2}
\end{table}

For what concerns the comparison with the state-of-the-art, results of DeepESN in this paper compare well with literature approaches recently devised on the same type of input data. In particular, in \cite{tablet} it is introduced a method for diagnosis of PD based on the analysis of spiral tests gathered from a similar tablet device, achieving a classification accuracy in PD identification of $79.1\%$. Interestingly, our approaches outperform such method obtaining
a test
accuracy of $84.13\%$, $87.20\%$ and $89.33\%$ achieved by shallowESN, DeepESN and ensemble of DeepESNs respectively.
This comparison further indicates that, contrary to what might appear at first glance from the example in 
Fig. \ref{fig:spiral_test},
simple statistics on the input signals, e.g. pressure data, are not rich enough to accurately discriminate PD\footnote{
Correlation 
between averaged pressure and PD in the dataset used in our paper is weak ($-0.31$).},
while our approach can capture relevant information from the whole temporal signal,
allowing to effectively improve such results.

\section{Conclusions}
In this paper, we proposed a novel approach for diagnosis of PD based on \mbox{DeepESN} model. 
The deep recurrent model is fed by the whole time-series gathered from a tablet during the sketching of spiral tests.
We performed a comparative assessment of our approach on a public dataset containing spiral tests executed by PD and control patients.
Results showed that the predictive accuracy obtained by \mbox{DeepESN}
significantly outperforms the result of shallow ESNs,
highlighting the potential benefits of 
deep recurrent architectures in the treatment of temporal signals for PD diagnosis.
Moreover, the use of ensemble method on the selected DeepESN model led to a further performance enhancement.
Overall, the proposed approach compared well also
with respect to state-of-the-art results,
further stressing the potentiality of exploiting the whole richness of temporal signals for PD diagnosis.

At the best of our knowledge, this work represents the first attempt to develop an approach for diagnosis of PD by using recurrent models, such as DeepESN, able to develop hierarchical temporal representations from tablet signals without the need of feature extraction and data preprocessing. 

\begin{footnotesize}

\bibliographystyle{unsrt}
\bibliography{references}

\end{footnotesize}

\end{document}